\documentclass[11pt]{article}
\usepackage{adjustbox}
\usepackage[final]{acl}
\usepackage{times}
\usepackage{graphicx}
\usepackage{subcaption}
\usepackage{latexsym}
\usepackage{booktabs}
\usepackage{multirow}
\usepackage{amssymb}
\usepackage{algorithm}

\usepackage{algpseudocode}
\usepackage{tabularx}
\usepackage[T1]{fontenc}
\usepackage{amsmath}
\usepackage[utf8]{inputenc}

\usepackage{microtype}

\usepackage{inconsolata}

\usepackage[table]{xcolor} 

\definecolor{toprow}{RGB}{232,241,250}
\definecolor{randrow}{RGB}{245,245,245}
\definecolor{colarbg}{RGB}{232,241,250}
\definecolor{LSTRbg}{RGB}{234,247,238}
\definecolor{graybg}{RGB}{242,242,242}
\usepackage{enumitem}

\definecolor{graybg}{gray}{0.9} 
\definecolor{cellblue}{HTML}{E6F0FF} 
\definecolor{cellgreen}{HTML}{E6FFE6} 

\newcommand{\ours}{LSTR}
\newcommand{\Deepseek}{\texttt{DeepSeek-R1-Distill-Qwen-7B}}
\newcommand{\Llama}{\texttt{Llama-3.2-8B-Instruct}}

\title{Beyond Dense States: Sparse Transcoders as Causally \\ Testable 
Operators for LLM Latent Reasoning}

\author{
  \textbf{Yadong Wang\textsuperscript{1,2}},
  \textbf{Haodong Chen\textsuperscript{1,2}},
  \textbf{Yu Tian\textsuperscript{3}}, \\
  \textbf{Chuanxing Geng\textsuperscript{1}},
  \textbf{Dong Liang\textsuperscript{1}},
  \textbf{Xiang Chen\textsuperscript{1,2}}\thanks{Corresponding author.} \\
  \textsuperscript{1}
  Nanjing University of Aeronautics and Astronautics \\
  \textsuperscript{2}State Key Laboratory of Ocean Sensing, 
ZJU-Hangzhou Global Scientific  and \\ Technological Innovation Center, 
Zhejiang University, Hangzhou, 311215, China \\
  \textsuperscript{3}360 Security Capability Center\\
 {  \texttt{\{adamwang373,xiang\_chen\}@nuaa.edu.cn}}\\
}

\begin{document}
\maketitle
\begin{abstract}
Latent reasoning reduces the token-generation cost of chain-of-thought reasoning by replacing explicit intermediate tokens with continuous latent transitions. However, existing latent reasoning methods usually rely on dense and entangled transitions, making their reasoning trajectories difficult to inspect or intervene on. We introduce \textbf{\ours} (\textbf{L}atent \textbf{S}parse \textbf{T}ranscoder \textbf{R}easoning), a framework that turns sparse transcoders from post-hoc diagnostic tools into in-loop, intervenable transition components for latent reasoning. At each latent step, a \textbf{Latent Transition Transcoder (LTT)} combines a linear skip path with a $\mathrm{Top}\text{-}k$ sparse innovation path, exposing a small set of active sparse features. Under matched compression settings, \ours{} offers a mechanistically inspectable alternative to dense latent reasoning. On GSM8K-Aug, ablating only a few top-active sparse features reduces accuracy by up to \textbf{$16.5\%$}, whereas analogous interventions have much smaller effects in dense latent baselines. These results indicate that the active sparse features are causally involved in the latent transition process, rather than merely post-hoc descriptors. Additional experiments on mathematical benchmarks and StrategyQA suggest that sparse latent transitions can preserve the compression benefits of latent reasoning while making the resulting trajectories more inspectable and intervenable.
\end{abstract}

\section{Introduction}

\begin{figure}[t]
    \centering
    \includegraphics[width=\columnwidth]{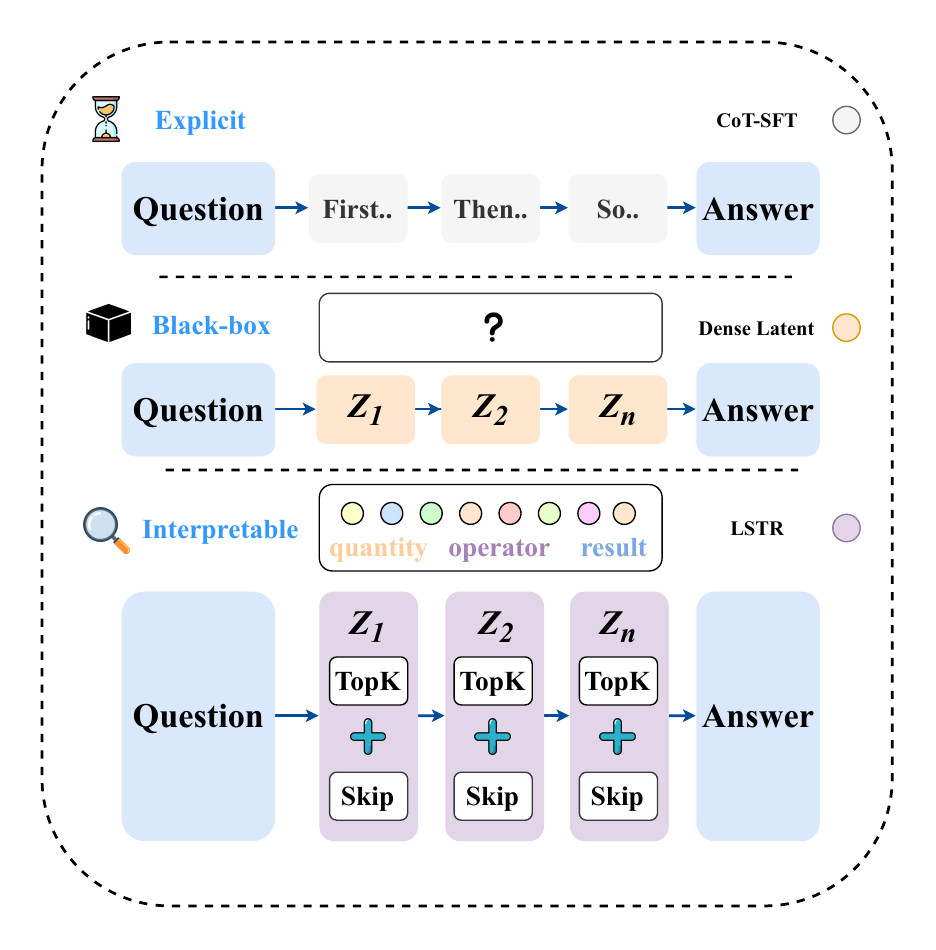}
    \caption{Comparison of reasoning paradigms. Explicit CoT uses long token sequences, while dense latent reasoning compresses intermediate computation into opaque continuous states. \ours{} computes each latent transition through a bounded set of $\mathrm{Top}\text{-}k$ sparse features, exposing transition-level units for inspection and intervention.}
    \label{fig:overview}
\end{figure}

Chain-of-Thought (CoT) prompting has substantially improved the reasoning capabilities of LLMs by decomposing complex tasks into explicit intermediate steps~\cite{DBLP:journals/corr/abs-2501-04682,DBLP:conf/nips/Wei0SBIXCLZ22,DBLP:journals/corr/abs-2412-16720,deco}. However, this paradigm introduces a significant computational bottleneck. Extended token-level reasoning chains increase inference latency and memory consumption, thereby restricting deployment in time-sensitive and resource-constrained environments. Because reasoning depth is tightly coupled with autoregressive token generation, the computational cost scales approximately linearly with the number of reasoning steps, rapidly becoming prohibitive for complex tasks.

To address this limitation, recent studies have explored latent reasoning, in which intermediate reasoning steps are compressed into continuous hidden representations rather than explicit language tokens~\cite{knowprompt,DBLP:conf/iclr/SaunshiDLKR25,DBLP:conf/acl/ChenSZXSLWSW025,11193717}. Methods such as Coconut~\cite{DBLP:journals/corr/abs-2412-06769} and CoLaR~\cite{DBLP:journals/corr/abs-2505-16552} demonstrate that multiple reasoning steps can be compressed into a small number of latent transitions, thereby substantially reducing overall inference costs. Despite these computational efficiency gains, existing latent reasoning methods remain fundamentally difficult to interpret. Most approaches rely on dense latent representations, in which all latent dimensions remain active throughout the entire reasoning process. Consequently, the internal reasoning dynamics become mechanistically opaque. Specifically, there is no principled mechanism for identifying which latent dimensions contribute to specific reasoning operations, a structural entanglement that severely limits the ability to inspect or intervene on latent reasoning trajectories.

In parallel, advances in \textbf{mechanistic interpretability} have shown that dense neural representations can often be decomposed into sparse sets of semantically meaningful features. Sparse autoencoders~\cite{bricken2023monosemanticity} and transcoder models~\cite{DBLP:conf/nips/DunefskyCN24,DBLP:journals/corr/abs-2501-18823} have uncovered interpretable semantic structures in frozen language models. Nevertheless, these methods remain primarily \textbf{post-hoc}. Sparse features are extracted from pretrained activations for diagnostic analysis, but they do not directly participate in the reasoning process during inference.

This separation exposes a fundamental gap between two research directions. Latent reasoning improves computational efficiency but sacrifices interpretability, whereas sparse feature learning improves interpretability while remaining disconnected from reasoning dynamics. This gap motivates a central question: \emph{can compressed latent reasoning be parameterized through a sparse transition interface that remains competitive with dense latent transitions while supporting direct feature-level intervention?}

This work provides an affirmative answer to this question by introducing \ours{} (Latent Sparse Transcoder Reasoning), a framework that reformulates latent reasoning as a sequence of transitions through a sparse feature space. In contrast to conventional dense latent reasoning, \ours{} imposes an explicit sparsity bottleneck at each latent step, ensuring that only a small subset of features is activated to contribute to each latent transition update. The core component of \ours{} is the LTT, which integrates a linear path with a sparse path. This design decouples background manifold transport from sparse residual updates. Consequently, sparsity in \ours{} is not merely a regularization strategy applied after the reasoning computation, but a fundamental architectural principle governing how reasoning computations are distributed and interpreted across latent dimensions.

A direct consequence of this formulation is \textbf{sparsity-mediated capacity control}. The sparsity budget $k$ determines the number of active sparse features during inference, thereby providing a mechanism for regulating transition capacity without retraining. More importantly, the sparse feature structure of \ours{} enables direct causal analysis of latent reasoning dynamics. Controlled ablation experiments demonstrate that removing a small number of highly activated sparse features substantially degrades reasoning performance in \ours{}, whereas the same intervention produces negligible effects in dense latent baselines. These findings indicate that the active sparse features are not merely diagnostic representations, but causally involved transition components.

Rather than treating latent reasoning only as a compression mechanism, we study whether its transition process can be given a sparse and causally testable interface. In this view, explicit CoT serves as an uncompressed reference, while the central comparison is whether sparse latent transitions can preserve the accuracy--length trade-off of dense latent reasoning while exposing feature-level intervention points. Experiments on mathematical reasoning benchmarks, including GSM8K-Aug, GSM-Hard, SVAMP, MultiArith, and MATH, as well as the commonsense reasoning benchmark StrategyQA, demonstrate that \ours{} maintains competitive performance under latent compression. Furthermore, the proposed approach exposes active sparse features that can be directly inspected, ablated, and causally tested. The primary contributions of this work are summarized as follows:
\begin{itemize}
    \item We introduce \textbf{\ours{}}, a sparse latent reasoning framework that uses in-loop transcoders as transition operators rather than post-hoc analysis modules.
    \item We provide quantitative causal evidence that a small set of active sparse features functionally supports latent reasoning, as targeted ablation produces substantially larger degradation than random or dense-state ablation.
    \item We show that the sparsity budget $k$ provides an inference-time mechanism for regulating latent transition capacity, enabling feature-level inspection and intervention without retraining.
\end{itemize}

\begin{figure*}[t]
    \centering
    \includegraphics[width=1.0\textwidth]{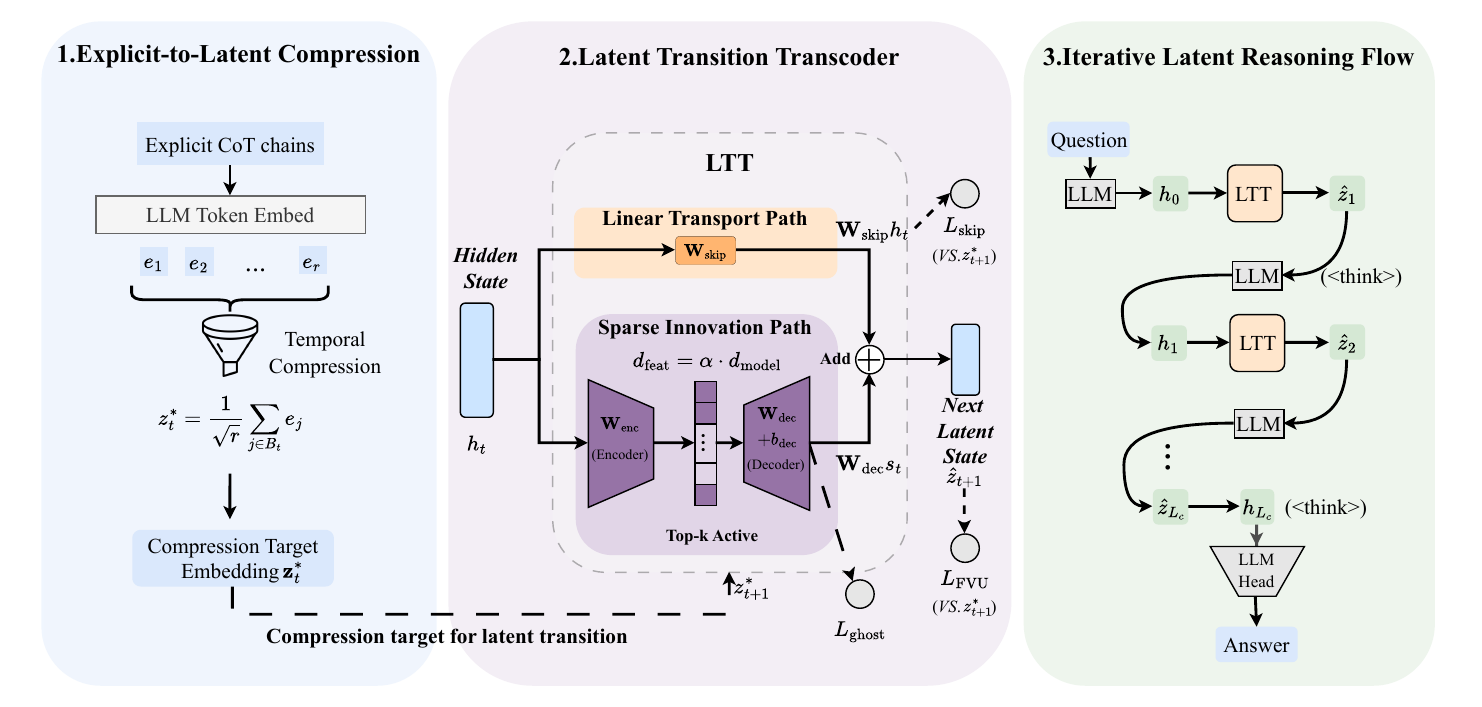}
    \caption{Overview of \ours{}. Contiguous explicit reasoning tokens are temporally aggregated into a compressed target embedding. At each latent step, the Latent Transition Transcoder (LTT) bilaterally decomposes transition dynamics: a linear Transport Path maintains manifold continuity, while a Sparse Innovation Path injects bounded sparse updates via $\mathrm{Top}\text{-}k$ selection. Under a fixed $(r,k)$ budget, LSTR exposes bounded active sparse features per step, facilitating inspection and intervention along the latent trajectory.}
    \label{fig:framework}
\end{figure*}

\section{Related Work}

\subsection{Latent and Implicit Reasoning}
Explicit CoT generation incurs substantial computational overhead and latency~\cite{DBLP:conf/nips/Wei0SBIXCLZ22}. To address this limitation, \textbf{latent reasoning} internalizes the reasoning process into continuous hidden representations~\cite{DBLP:journals/corr/abs-2507-06203}. While early methods such as iCoT~\cite{DBLP:journals/corr/abs-2405-14838} absorb reasoning supervision directly into model parameters, recent approaches explicitly model latent trajectories. For example, Coconut~\cite{DBLP:journals/corr/abs-2412-06769} introduces a ``chain of continuous thought'' by recursively feeding hidden states back into the model, thereby enabling exploration across latent reasoning paths. Furthermore, CODI~\cite{DBLP:journals/corr/abs-2502-21074} aligns the latent states of a student model with the reasoning hidden states of a teacher model via self-distillation. CoLaR~\cite{DBLP:journals/corr/abs-2505-16552} subsequently implements dynamic compression and optimizes these trajectories using reinforcement learning. Despite their effectiveness in reducing inference costs, these approaches rely almost exclusively on \textbf{dense} latent representations, wherein all latent dimensions remain jointly active during each transition. This dense formulation inherently limits interpretability and controllability, as individual latent dimensions lack clear semantic attribution. In contrast, the proposed framework introduces an explicit sparse semantic bottleneck, enabling latent transitions to be composed of localized and interpretable functional features.

\subsection{Sparse Transcoders and Feature Learning}

Sparse Autoencoders (SAEs) disentangle polysemantic neurons into 
interpretable semantic features~\cite{bricken2023monosemanticity}. 
More recently, \textbf{Transcoders} extend this paradigm by modeling 
the functional input--output behavior rather than reconstructing 
static activations~\cite{DBLP:conf/nips/DunefskyCN24}. Empirical 
studies show that transcoders capture substantially more interpretable 
features by directly approximating the operational semantics of neural 
submodules~\cite{DBLP:journals/corr/abs-2501-18823}. Architectural 
refinements such as affine residual paths alleviate the rank 
limitations imposed by sparse bottlenecks, while ghost-gradient updates 
revive underutilized features by propagating residual signals to 
inactive units. However, existing sparse feature learning remains 
primarily a post-hoc diagnostic tool for static and frozen models. In 
contrast, our work repurposes sparse transcoders as \emph{active 
reasoning operators}, with sparse features participating directly in 
latent state transitions rather than describing them post-hoc. The key distinction is not merely that LSTR produces sparse activations, but that these activations are part of the state transition that constructs the next latent reasoning state.

\section{Methodology}
\label{sec:method}

\begin{table*}[t]
\centering
\small
\renewcommand{\arraystretch}{1.1}
\setlength{\tabcolsep}{4pt}
\caption{Performance comparison on four grade-school mathematical reasoning benchmarks. We report averaged accuracy (Acc. \%) and average latent reasoning length (\#L) over five runs with different random seeds, together with 95\% confidence intervals (±). Gray rows indicate ablation variants of \ours{}. Among non-CoT methods, bold marks the best average accuracy, while underlining highlights a nearly matched result.}
\label{tab:results}
\begin{tabular}{lcccccccccc}
\toprule
\multirow{2}{*}{\textbf{Method}} 
& \multicolumn{2}{c}{\textbf{GSM8K-Aug}} 
& \multicolumn{2}{c}{\textbf{GSM-Hard}} 
& \multicolumn{2}{c}{\textbf{SVAMP}} 
& \multicolumn{2}{c}{\textbf{MultiArith}} 
& \multicolumn{2}{c}{\textbf{Average}} \\
& Acc. & \#L & Acc. & \#L & Acc. & \#L & Acc. & \#L & Acc. & \#L \\
\midrule
CoT 
& $49.4_{\pm.72}$ & $25.6_{\pm.11}$ 
& $11.9_{\pm.16}$ & $34.2_{\pm.11}$ 
& $59.8_{\pm.29}$ & $12.1_{\pm.03}$ 
& $93.2_{\pm.49}$ & $13.7_{\pm.09}$ 
& 53.6 & 21.4 \\
\midrule
iCoT 
& $19.8_{\pm.23}$ & $0.00_{\pm.00}$ 
& $3.87_{\pm.16}$ & $0.00_{\pm.00}$ 
& $36.4_{\pm.51}$ & $0.00_{\pm.00}$ 
& $38.2_{\pm.66}$ & $0.00_{\pm.00}$ 
& 24.6 & 0.00 \\

Coconut 
& $23.1_{\pm.28}$ & $6.00_{\pm.00}$ 
& $5.49_{\pm.33}$ & $6.00_{\pm.00}$ 
& $40.7_{\pm.65}$ & $6.00_{\pm.00}$ 
& $41.1_{\pm.24}$ & $6.00_{\pm.00}$ 
& 27.6 & 6.00 \\

Distill 
& $13.3_{\pm.62}$ & $6.00_{\pm.00}$ 
& $2.97_{\pm.24}$ & $6.00_{\pm.00}$ 
& $21.7_{\pm.73}$ & $6.00_{\pm.00}$ 
& $19.2_{\pm.83}$ & $6.00_{\pm.00}$ 
& 14.3 & 6.00 \\

\midrule

CoLaR-5 
& $26.8_{\pm.17}$ & $5.57_{\pm.02}$ 
& $5.87_{\pm.10}$ & $6.53_{\pm.01}$ 
& $48.4_{\pm.45}$ & $2.95_{\pm.02}$ 
& $86.4_{\pm.35}$ & $3.21_{\pm.01}$ 
& 41.7 & 4.57 \\

\ours-5 
& $29.5_{\pm.32}$ & $5.69_{\pm.00}$ 
& $6.29_{\pm.12}$ & $6.75_{\pm.00}$ 
& $49.2_{\pm.49}$ & $3.03_{\pm.00}$ 
& $88.2_{\pm.14}$ & $3.17_{\pm.00}$ 
& \textbf{43.3} & 4.66 \\

\rowcolor{graybg}
\quad - w/o Skip 
& $27.5_{\pm.09}$ & $5.75_{\pm.00}$ 
& $5.52_{\pm.16}$ & $7.95_{\pm.00}$ 
& $50.5_{\pm.59}$ & $3.18_{\pm.00}$ 
& $84.1_{\pm.32}$ & $3.23_{\pm.00}$ 
& 41.9 & 5.0 \\

\rowcolor{graybg}
\quad - w/o Sparse 
& $20.7_{\pm.29}$ & $5.80_{\pm.00}$ 
& $4.44_{\pm.11}$ & $6.58_{\pm.00}$ 
& $41.7_{\pm.48}$ & $3.11_{\pm.00}$ 
& $75.8_{\pm.29}$ & $3.38_{\pm.00}$ 
& 35.6 & 4.7 \\

CoLaR-2 
& $40.1_{\pm.20}$ & $12.7_{\pm.02}$ 
& $9.08_{\pm.03}$ & $14.0_{\pm.07}$ 
& $54.9_{\pm.20}$ & $6.11_{\pm.01}$ 
& $91.3_{\pm.12}$ & $7.35_{\pm.01}$ 
& \textbf{48.8} & 10.0 \\

\ours-2 
& $40.2_{\pm.09}$ & $13.18_{\pm.00}$ 
& $8.81_{\pm.06}$ & $18.51_{\pm.00}$ 
& $53.9_{\pm.29}$ & $6.47_{\pm.00}$ 
& $90.5_{\pm.29}$ & $7.44_{\pm.00}$ 
& \underline{48.4} & 11.4 \\

\rowcolor{graybg}
\quad - w/o Skip 
& $35.3_{\pm.19}$ & $13.14_{\pm.00}$ 
& $7.90_{\pm.12}$ & $24.41_{\pm.00}$ 
& $53.1_{\pm.29}$ & $6.69_{\pm.00}$ 
& $92.8_{\pm.21}$ & $7.37_{\pm.00}$ 
& 47.3 & 12.9 \\

\rowcolor{graybg}
\quad - w/o Sparse 
& $34.2_{\pm.22}$ & $13.43_{\pm.00}$ 
& $7.32_{\pm.17}$ & $23.37_{\pm.00}$ 
& $48.6_{\pm.36}$ & $6.49_{\pm.00}$ 
& $85.3_{\pm.37}$ & $7.57_{\pm.00}$ 
& 43.9 & 12.7 \\
\bottomrule
\end{tabular}
\end{table*}

\subsection{Overview of LSTR}
\label{sec:method_overview}

We introduce \ours{} (Latent Sparse Transcoder Reasoning), a framework parameterizing latent reasoning as a sequence of sparse transitions. Existing latent reasoning methods replace explicit CoT tokens with continuous latent transitions, but these transitions are typically dense: all dimensions remain active, making individual component contributions difficult to inspect or intervene on. \ours{} retains the compressed latent setup but redesigns the transition function. At each latent step, the model computes the subsequent continuous state using a small set of active sparse features.

Given a backbone hidden state, the core LTT module predicts the next latent state via a linear skip path and a sparse innovation path. The skip path facilitates the linear transport of predictable variations between consecutive states, whereas the sparse path models residual nonlinear updates using a bounded set of dictionary features. Rather than assuming perfect separation, this architectural inductive bias establishes a sparse transition interface that facilitates evaluation through feature ablation and intervention. Unlike post-hoc feature extraction, \ours{} generates sparse activations internally within the latent reasoning loop to construct the subsequent state directly. Furthermore, the sparsity budget $k$ constrains the number of active features per step, thereby enabling capacity control at inference time without requiring retraining (Fig.~\ref{fig:framework}). LSTR provides an architectural substrate for interpretability by exposing bounded active features; the semantic mapping of these features remains an open empirical question that our framework enables but does not resolve.

\subsection{Latent Trajectory Construction and Inference}
\label{sec:latent_trajectory}

Given an explicit reasoning sequence $T_r$ of length $L_r$, \ours{} maps it to a compressed latent trajectory $\mathcal{Z}^*=\{\mathbf{z}_1^*,\dots,\mathbf{z}_{L_c}^*\}$ with a compression ratio $r$, where $L_c=\lceil L_r/r\rceil$. Each latent target is derived by aggregating a contiguous block of token embeddings $\mathcal{B}_t$:
\begin{equation}
\mathbf{z}_t^{\star}
=
\frac{1}{\sqrt{r}}
\sum_{j\in \mathcal{B}_t}\mathbf{e}_j .
\end{equation}
The $1/\sqrt{r}$ scaling factor stabilizes the variance as $r$ varies, preventing the latent magnitude from scaling with the block size. To ensure numerical stability, the latent targets are further standardized using empirical embedding statistics.

During training, we optimize over the concatenated sequence $[\mathbf{E}_q,\text{\texttt{<think>}},\mathcal{Z}^*,\text{\texttt{<think>}},\mathbf{E}_a]$. At latent step $t$, the LTT predicts $\hat{\mathbf{z}}_{t+1}$ from the backbone hidden state $\mathbf{h}_t$, and this prediction is supervised by $\mathbf{z}_{t+1}^*$. Concurrently, the language modeling head predicts a randomly sampled token from the corresponding block $\mathcal{B}_{t+1}$. This auxiliary token objective anchors the latent transition to discrete language semantics without requiring complete token-level reasoning generation. Furthermore, random sampling within each block mitigates positional bias.

At inference time, explicit intermediate reasoning tokens are replaced by iterative continuous transitions. Given a question prefix followed by \texttt{<think>}, the backbone generates an initial hidden state $\mathbf{h}_0$. The LTT maps $\mathbf{h}_0$ to a sparse activation vector $\mathbf{s}_0$ and a latent state $\hat{\mathbf{z}}_1$. This latent state is then fed back into the backbone as a continuous reasoning input to produce the subsequent hidden state $\mathbf{h}_1$. This procedure is repeated for $L_c$ latent steps. Finally, the model exits the latent reasoning segment and generates the final answer using the language modeling head. Formally, this iterative process is defined as:
\begin{equation}
\begin{aligned}
\mathbf{h}_t
&=
f_{\theta}(\mathbf{x}_{\leq t}),\\
(\mathbf{s}_t,\hat{\mathbf{z}}_{t+1})
&=
\mathrm{LTT}(\mathbf{h}_t),\\
\mathbf{x}^{\mathrm{lat}}_{t+1}
&\leftarrow
\hat{\mathbf{z}}_{t+1},
\end{aligned}
\end{equation}
where $f_{\theta}$ denotes the backbone language model and $\mathbf{x}^{\mathrm{lat}}_{t+1}$ represents the continuous latent input consumed at the next latent step.

\subsection{Latent Transition Transcoder}
\label{sec:ltt}

The \textbf{LTT} predicts the next latent state $\hat{\mathbf{z}}_{t+1}$ from the backbone hidden state $\mathbf{h}_t\in\mathbb{R}^{d_{\text{model}}}$:
\begin{equation}
\hat{\mathbf{z}}_{t+1}
=
\mathbf{W}_{\mathrm{skip}}\mathbf{h}_t
+
\mathbf{W}_{\mathrm{dec}}\mathbf{s}_t
+
\mathbf{b}_{\mathrm{dec}},
\end{equation}
\begin{equation}
\mathbf{s}_t
=
\mathrm{Top}\text{-}k
\left(
\sigma\left(
\mathbf{W}_{\mathrm{enc}}(\mathbf{h}_t-\boldsymbol{\mu})
+
\mathbf{b}_{\mathrm{enc}}
\right)
\right).
\end{equation}

Here, $\mathbf{W}_{\mathrm{skip}}\in\mathbb{R}^{d_{\text{model}}\times d_{\text{model}}}$ is a zero-initialized, bias-free linear adapter. The sparse path projects $\mathbf{h}_t$ into an overcomplete dictionary with $d_{\mathrm{feat}}=\alpha d_{\text{model}}$, where $\alpha$ is the expansion factor, and $\sigma(\cdot) = \mathrm{ReLU}(\cdot)$ is the activation function. The $\mathrm{Top}\text{-}k$ operator retains the $k$ largest activations and sets the remainder to zero, yielding a sparse vector $\mathbf{s}_t\in\mathbb{R}^{d_{\mathrm{feat}}}$.

Several design choices are intended to enhance stability under iterative latent supervision. Empirical mean subtraction with $\boldsymbol{\mu}$ eliminates dominant global activation directions that would otherwise occupy sparse slots across multiple steps. Decoder columns are constrained to unit norm, $\|\mathbf{W}_{\mathrm{dec}}^{(i)}\|_2=1$, to prevent feature-scale collapse. Hard $\mathrm{Top}\text{-}k$ selection maintains a fixed number of active features, thereby avoiding uncontrolled variations in transition capacity along the trajectory. Gradients through the non-differentiable $\mathrm{Top}\text{-}k$ selection are approximated using a Straight-Through Estimator, allowing selected sparse features to be optimized under hard sparsity constraints. Inactive or under-utilized features are further addressed by the feature-utilization objective detailed below.

\subsection{Optimization and Feature Utilization}
\label{sec:optimization}

The complete training loss combines language-model supervision with latent transition supervision:
\begin{equation}
\mathcal{L}
=
\mathcal{L}_{\mathrm{LM}}
+
\beta\mathcal{L}_{\mathrm{latent}},
\end{equation}
where $\mathcal{L}_{\mathrm{LM}}$ is the cross-entropy loss applied over answer tokens and randomly sampled block-level token targets. The latent objective is formulated as:
\begin{equation}
\mathcal{L}_{\mathrm{latent}}
=
\mathcal{L}_{\mathrm{FVU}}
+
\lambda_s\mathcal{L}_{\mathrm{skip}}
+
\lambda_g\mathcal{L}_{\mathrm{ghost}}.
\end{equation}

The primary reconstruction term utilizes the \textbf{Fraction of Variance Unexplained (FVU)} rather than raw squared error:
\begin{equation}
\mathcal{L}_{\mathrm{FVU}}
=
\frac{
\mathbb{E}
\left\|
\hat{\mathbf{z}}_{t+1}
-
\mathbf{z}_{t+1}^*
\right\|_2^2
}{
\mathrm{Var}(\mathbf{z}_{t+1}^*)
}.
\end{equation}
This normalization accounts for variations in latent magnitude across reasoning depths and compression ratios, preventing high-magnitude targets from dominating the optimization process.

The skip objective introduces an auxiliary bias for the skip path:
\begin{equation}
\mathcal{L}_{\mathrm{skip}}
=
\mathbb{E}
\left\|
(
\mathbf{W}_{\mathrm{skip}}\mathbf{h}_t
+
\mathbf{b}_{\mathrm{dec}}
)
-
\mathbf{z}_{t+1}^*
\right\|_2^2 .
\end{equation}
Because the skip path is linear and bias-free, it is ideally suited to capture predictable components of consecutive latent states. This reduces the representational burden on the sparse path, whose active capacity is inherently restricted by the $\mathrm{Top}\text{-}k$ selection.

Under hard sparsity, certain encoder units may rarely enter the active set. To improve dictionary utilization, \ours{} incorporates a ghost-gradient objective that propagates reconstruction residuals to inactive or under-utilized encoder features. Let $\mathbf{r}_t=\mathbf{z}_{t+1}^*-\hat{\mathbf{z}}_{t+1}$ denote the transition residual. The ghost objective is defined as:
\begin{equation}
\mathcal{L}_{\mathrm{ghost}}
=
\mathbb{E}
\left\|
\mathbf{W}_{\mathrm{dec}}
\,
\mathrm{enc}(\mathbf{h}_t)_{\mathrm{dead}}
-
\mathbf{r}_t
\right\|_2^2 .
\end{equation}
This auxiliary term encourages unused dictionary features to align with unexplained residual directions, thereby mitigating feature collapse during training.

\section{Experiments}
\label{sec:experiments}

\begin{figure*}[t]
    \centering
    \includegraphics[width=0.98\linewidth]{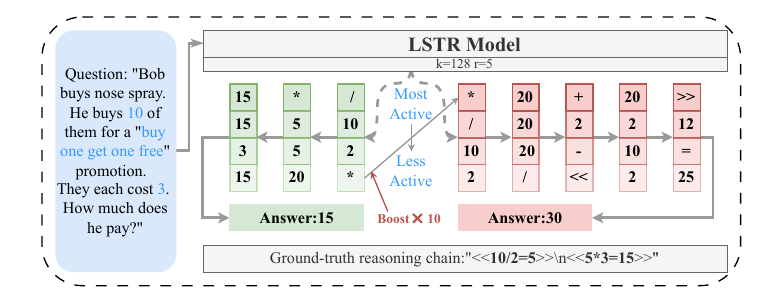}
    \caption{A qualitative illustration of how perturbing a single active feature can alter the latent trajectory. Systematic semantic annotation of sparse features is left to future work.q}
    \label{fig:case_study}
\end{figure*}

Our evaluation of \ours{} begins by testing whether sparse transitions maintain the trade-off between accuracy and trajectory length in latent reasoning under matched compression. We then examine the model's internal dynamics during inference, specifically the functional roles of the two LTT pathways and the active sparse features. We also analyze how the sparsity budget controls transition capacity and whether these behaviors generalize beyond arithmetic tasks. Throughout the experiments, explicit CoT serves as an uncompressed reference, while dense latent reasoning methods, particularly CoLaR, act as the primary baseline under identical compression conditions.

\begin{table}[t]
\centering
\small
\setlength{\tabcolsep}{6pt}
\renewcommand{\arraystretch}{1.12}
\caption{Inference-time path masking on GSM8K. All rows use the same trained \ours{} model.}
\label{tab:pathway_masking}
\begin{tabular}{llcc}
\toprule
Masked path & Remaining paths & Acc. (\%) & $\Delta$ \\
\midrule
\rowcolor{gray!10}
None & Skip + Sparse & \textbf{40.2} & -- \\
Sparse & Skip only & 35.2 & -5.0 \\
Skip & Sparse only & 0.3 & -39.9 \\
\bottomrule
\end{tabular}
\end{table}

\subsection{Experimental Setup}
\label{sec:exp_setup}

\paragraph{Datasets and metrics.}
We primarily train and evaluate our models on \textbf{GSM8k-Aug}~\cite{DBLP:journals/corr/abs-2311-01460}, an augmented version of GSM8k containing approximately 385k training samples and 1k test samples.
To assess generalization across reasoning distributions, we additionally evaluate on three out-of-domain benchmarks:
\textbf{GSM-Hard}~\cite{DBLP:conf/icml/GaoMZ00YCN23},
\textbf{SVAMP}~\cite{DBLP:conf/naacl/PatelBG21}, and
\textbf{MultiArith}~\cite{DBLP:journals/corr/RoyR16}.
For evaluating scalability to more challenging reasoning regimes, we further report results on the \textbf{MATH} dataset~\cite{DBLP:conf/nips/HendrycksBKABTS21}. Following prior work~\cite{DBLP:journals/corr/abs-2412-06769}, we report
\textbf{Accuracy (Acc.)} as the percentage of correctly solved problems and
\textbf{Reasoning Length (\#L)}, measured as the average number of generated reasoning tokens or latent transitions, as a proxy for inference cost.

\paragraph{Baselines.}
We compare \ours~against the following representative baselines:
(1) \textbf{CoT}~\cite{DBLP:conf/nips/Wei0SBIXCLZ22}, fine-tuned on complete reasoning chains, performs token-by-token reasoning before outputting answers;
(2) \textbf{iCoT}~\cite{DBLP:journals/corr/abs-2405-14838}, gradually internalizes reasoning steps and directly generates answers without intermediate tokens;
(3) \textbf{Coconut}~\cite{DBLP:journals/corr/abs-2412-06769}, progressively replaces token reasoning with latent steps through curriculum learning, executing fixed-length latent reasoning;
(4) \textbf{Distill}~\cite{DBLP:journals/corr/abs-2502-21074}, distills token-level CoT into latent reasoning steps and infers similarly to Coconut.
(5) \textbf{CoLaR}~\cite{DBLP:journals/corr/abs-2505-16552}, performs dynamic latent reasoning with compression, trained for speed-adjustable inference.

\paragraph{Implementation details.}
All models use the \texttt{Llama-3.2-1B-Instruct} backbone~\cite{DBLP:journals/corr/abs-2407-21783}. We freeze the backbone parameters and employ \textbf{LoRA} for parameter-efficient fine-tuning. Following~\cite{DBLP:journals/corr/abs-2505-16552}, each model is trained for at most 50 epochs or 12 hours. The checkpoint with the best validation accuracy is used for evaluation. For \ours{}, the feature expansion factor is set to 16, and the default sparsity budget is $k=128$ active features per latent step. Unless otherwise stated, results are averaged over five random seeds. Appendix~\ref{app:impl} gives the full hyperparameters, optimizer settings, and hardware details.

\subsection{Performance under Latent Compression}
\label{sec:main_results}

Table~\ref{tab:results} presents the main results across four mathematical reasoning benchmarks. The objective of this comparison is not to demonstrate that latent reasoning achieves higher accuracy than explicit CoT. Explicit CoT remains the strongest token-level reference across several benchmarks, despite requiring substantially longer reasoning chains. Instead, the primary comparison focuses on dense latent reasoning methods under identical compression settings, particularly CoLaR.

At strong compression ($r=5$), \ours{} outperforms CoLaR on GSM8K-Aug and GSM-Hard while performing comparably on SVAMP and MultiArith, increasing average accuracy from 41.7\% to 43.3\% at similar latent lengths. At milder compression ($r=2$), \ours{} averages 48.4\% accuracy versus 48.8\% for CoLaR, demonstrating sparse transitions competitively replace dense ones without compromising the accuracy-length trade-off (Appendix~\ref{app:r_sensitivity} details sensitivity for $r\in\{1,2,3,4,5\}$). Train-time ablations (Table~\ref{tab:results}, gray rows) reveal both pathways are essential when training from scratch, complementing Section~\ref{sec:causal_analysis}. Eliminating the sparse path degrades latent transition modeling, dropping accuracy from 43.3\% to 35.6\% at $r=5$ and 48.4\% to 43.9\% at $r=2$. Removing the skip path similarly reduces performance and extends trajectory length at $r=2$ from 11.4 to 12.9, confirming its stabilizing transport function.

\subsection{Causal Analysis of Sparse Transitions}
\label{sec:causal_analysis}

We next investigate whether \ours{} actively uses sparse transitions during inference, beyond merely altering model parameterization. We apply two interventions: masking both LTT paths in the same trained model to isolate their roles, and removing active sparse features to evaluate their impact on final reasoning outcomes.

\paragraph{Inference-time path masking.}
The LTT consists of a skip path and a sparse path. To isolate their respective effects, we train the complete \ours{} model once and mask one of the pathways during inference. Table~\ref{tab:pathway_masking} presents the evaluation results on GSM8K-Aug. The two pathways exhibit distinct failure modes. Removing the sparse path yields a skip-only model with an accuracy of 35.2\%, suggesting that the linear path sustains a substantial portion of the continuous latent trajectory. The performance gap of 4.9\% relative to the full model demonstrates that the sparse path provides critical residual updates beyond linear transport. Conversely, ablating the skip path reduces the accuracy to 0.3\%, indicating that sparse updates alone cannot maintain the latent trajectory in the absence of a stable transport channel. These findings highlight a clear functional division: the skip path preserves the continuity of the latent state, whereas the sparse path provides residual transition updates that enhance reasoning capabilities.

\begin{figure}[t]
    \centering
    \includegraphics[width=0.95\linewidth]{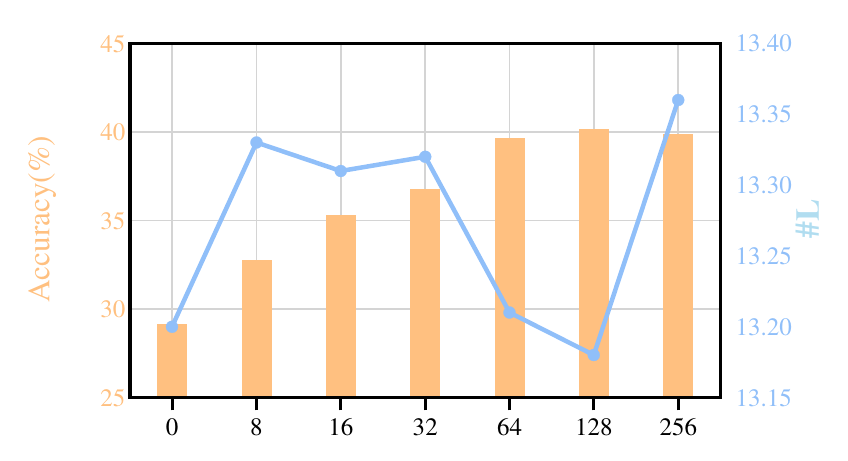}
    \caption{Accuracy and average latent reasoning length under different inference-time sparsity budgets $k$. The model is trained with $r=2$ and $k=128$, and evaluated by changing $k$ only at inference time.}
    \label{fig:k_sweep}
\end{figure}

\paragraph{Feature-level causal ablation.}
To evaluate whether active sparse features affect accuracy, we ablate the $N$ most active features per latent step and measure the accuracy change, comparing against a baseline of $N$ randomly ablated active features. For CoLaR, we identically ablate the $N$ latent dimensions with the largest absolute activations (Table~\ref{tab:causal_ablation}). Ablating the top-active features in \ours{} steadily degrades accuracy, dropping 16.5 percentage points at $N=10$, whereas ablating random active features yields a substantially smaller effect. Similar intervention on the top dense dimensions of CoLaR induces a smaller drop, confirming that computation is more distributed across dense latent states. While not implying every sparse feature maps to a fully interpretable concept, these results substantiate that active sparse features in \ours{} causally participate in the latent transition process.

\begin{table}[t]
\centering
\small
\setlength{\tabcolsep}{5pt}
\renewcommand{\arraystretch}{1.12}
\caption{Feature-level causal ablation on GSM8K-Aug. Accuracy change (percentage points) after removing $N$ sparse features or dense latent dimensions during inference. Blue and gray rows denote top-ranked and random units, respectively.}
\label{tab:causal_ablation}
\begin{tabular}{lrrrr}
\toprule
Model & $N=1$ & $N=3$ & $N=5$ & $N=10$ \\
\midrule
\rowcolor{toprow}
\ours{} & -4.6 & -11.8 & -15.1 & \textbf{-16.5} \\
\rowcolor{toprow}
CoLaR & +1.0 & +0.7 & -0.7 & -3.0 \\
\rowcolor{randrow}
\ours{} & -0.3 & -0.7 & -2.0 & -0.7 \\
\rowcolor{randrow}
CoLaR & +1.3 & -1.0 & -1.0 & +2.0 \\
\bottomrule
\end{tabular}
\end{table}

\paragraph{Qualitative feature intervention.}
For visualization, we assign feature labels by projecting sparse features onto the token embedding space with cosine similarity. This projection is only a weak semantic probe that helps make trajectories readable. The causal claim is based on Table~\ref{tab:causal_ablation}. Figure~\ref{fig:case_study} shows one intervention example, where modifying an early active feature changes the later trajectory and the final answer. Additional structural analyses are provided in Appendix~\ref{app:latent_structure}.

\subsection{Capacity Control and Sparse Dynamics}
\label{sec:capacity_dynamics}

We study how sparse transition capacity is controlled and used across latent reasoning trajectories. The sparsity budget $k$ sets the maximum number of active features at each latent step, allowing inference-time capacity adjustments without retraining. Figure~\ref{fig:k_sweep} shows that accuracy improves when more sparse features participate, while the average latent reasoning length remains relatively stable. This result indicates that $k$ directly controls the capacity of sparse latent transitions, rather than serving only as a post-hoc analysis parameter. This control should not be interpreted as semantic control over the model, since changing $k$ adjusts how many sparse features contribute to reasoning without determining which semantic operation the model performs. Appendix~\ref{app:r_sensitivity} reports a separate compression-ratio analysis, where $r$ controls latent trajectory length rather than the number of active features per transition.

\begin{figure}[t]
    \centering
    \includegraphics[width=0.95\linewidth]{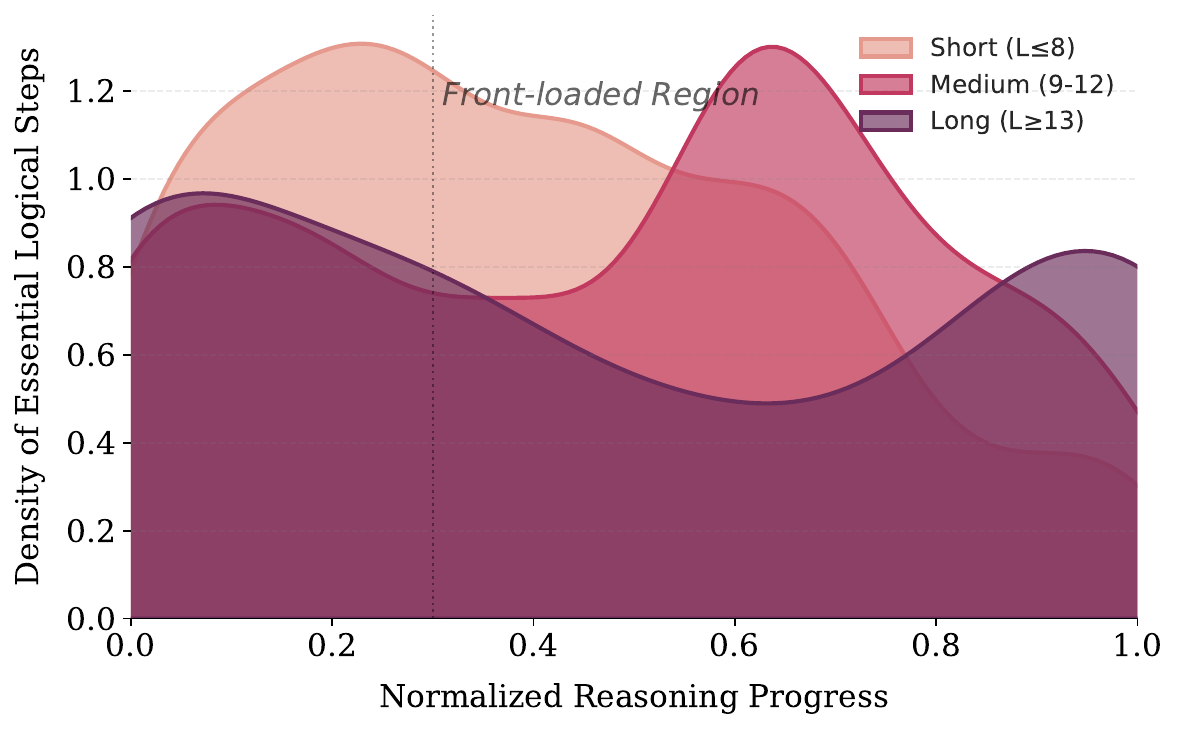}
    \caption{KDE-estimated temporal density of causally necessary latent steps ($r=2$) across heterogeneous trajectory lengths ($L \in [4, 35]$). Length-balanced cohorts exhibit a consistent front-loaded profile.}
    \label{fig:stepwise_ablation}
\end{figure}

We then examine where causal necessity manifests within compressed latent trajectories. Figure~\ref{fig:stepwise_ablation} estimates the distribution of causally necessary latent steps under $r=2$ across trajectories of different lengths. The density is consistently front-loaded across length-balanced cohorts, suggesting that early latent transitions carry a larger share of task-relevant updates. This pattern agrees with the feature-level ablation in Table~\ref{tab:causal_ablation}: sparse features are causally relevant in aggregate, and their influence is organized over time.

\subsection{Generalization and Efficiency}
\label{sec:generalization_tradeoff}

The main experiments focus on mathematical reasoning, where reasoning steps are closely tied to numbers and operators. 
To test whether sparse latent transitions also apply outside arithmetic, we evaluate \ours{} on StrategyQA~\cite{DBLP:journals/tacl/GevaKSKRB21}, a commonsense reasoning benchmark. 
Under the same compression setting, \ours{} achieves 61.13\% accuracy, compared with 58.49\% for CoLaR and 64.83\% for explicit CoT. 
This result suggests that sparse latent transitions are not limited to arithmetic reasoning, although evaluation on more natural-language reasoning tasks is still needed.

Appendix~\ref{app:overhead} analyzes computational overhead. 
The sparse transition operator adds per-step computation because it uses an expanded dictionary and Top-$k$ selection. 
On GSM8K-Aug, \ours{} ($r=2$) has higher per-step latency than CoLaR ($r=2$), but it reduces end-to-end time relative to explicit CoT by using a shorter reasoning trajectory. 
The efficiency gain of \ours{} therefore comes mainly from trajectory compression, not from faster individual latent transitions. 
Appendix~\ref{app:harder_tasks} reports results on harder tasks and a larger backbone, where the accuracy gap between latent reasoning and explicit CoT remains a major limitation on challenging mathematical benchmarks.

\section{Conclusion}

We propose \ours{}, advancing sparse transcoders from post-hoc diagnostics 
to direct transition operators in latent reasoning. It computes 
transitions using bounded $\mathrm{Top}\text{-}k$ sparse features and a 
linear skip path for predictable drift, matching dense baselines under 
matched compression while exposing feature-level trajectory access. 
Causal ablations show that removing top-active features substantially 
degrades the model but only weakly affects dense baselines, indicating 
these features causally participate in transitions. Although a gap 
with explicit Chain-of-Thought persists, \ours{} is a step toward 
causally testable reasoning. Future work includes reducing CoT 
supervision and developing stronger interpretation tools.

\section*{Limitations}

\ours{} remains subject to several limitations. First, it relies on compressed targets derived from explicit Chain-of-Thought traces, so it currently transfers existing reasoning structures into latent space rather than learning them without CoT supervision. Second, latent compression still trails explicit CoT on harder symbolic benchmarks such as MATH, where fine-grained intermediate states are important for long-horizon compositional reasoning. Third, the interpretability of \ours{} is local: feature-level interventions reveal causally active transition components, but they do not provide a complete symbolic reconstruction of the full reasoning trajectory. Finally, although we include StrategyQA as an initial non-arithmetic benchmark, most mechanistic analyses are still performed on mathematical tasks. Extending the evaluation to broader natural-language reasoning and agentic settings remains future work.

\section*{Ethics Statement}

This work utilizes publicly available reasoning benchmarks and pretrained language models, without involving new data collection, human-subject experiments, or the processing of personally identifiable information. The proposed sparse-transition interface is designed to support the debugging, mechanistic analysis, and auditing of latent reasoning. However, this capability for feature-level intervention may also facilitate more precise activation steering if subjected to misuse. Therefore, we present these interventions strictly for controlled analysis and emphasize that the application of such techniques in safety-critical or user-facing systems requires additional safeguards and continuous monitoring.

\section*{Acknowledgments}
 
This work was supported by the National Natural Science Foundation of China (Nos. 62506166, U2441285), the Natural Science Foundation
of Jiangsu Province (No. BK20251365), the China Postdoctoral Science Foundation (No. 2025M774283), 
the Scientific Research Starting Foundation of
Nanjing University of Aeronautics and Astronautics (No. 1015-YAH24096), and the Open Project of the State Key
Laboratory of Ocean Sensing (No. OSKF-2025M10).
This work was sponsored by the Chinese Association for Artificial Intelligence - Huawei AI Computing Power Acceleration Program.
This research was also supported by the DiDi GAIA Collaborative Research Funds (No. CCF-DiDi GAIA202507) and CAAI-MindSpore Open Fund (CAAIXSJLJJ 2025 MindSpore 01), developed on OpenI Community.

\bibliography{custom}

@inproceedings{DBLP:conf/icml/GaoMZ00YCN23,
  author       = {Luyu Gao and
                  Aman Madaan and
                  Shuyan Zhou and
                  Uri Alon and
                  Pengfei Liu and
                  Yiming Yang and
                  Jamie Callan and
                  Graham Neubig},
  editor       = {Andreas Krause and
                  Emma Brunskill and
                  Kyunghyun Cho and
                  Barbara Engelhardt and
                  Sivan Sabato and
                  Jonathan Scarlett},
  title        = {{PAL:} Program-aided Language Models},
  booktitle    = {International Conference on Machine Learning, {ICML} 2023, 23-29 July
                  2023, Honolulu, Hawaii, {USA}},
  series       = {Proceedings of Machine Learning Research},
  volume       = {202},
  pages        = {10764--10799},
  publisher    = {{PMLR}},
  year         = {2023},
  url          = {https://proceedings.mlr.press/v202/gao23f.html},
  bibsource    = {dblp computer science bibliography, https://dblp.org}
}

@inproceedings{DBLP:conf/iclr/SaunshiDLKR25,
  author       = {Nikunj Saunshi and
                  Nishanth Dikkala and
                  Zhiyuan Li and
                  Sanjiv Kumar and
                  Sashank J. Reddi},
  title        = {Reasoning with Latent Thoughts: On the Power of Looped Transformers},
  booktitle    = {The Thirteenth International Conference on Learning Representations,
                  {ICLR} 2025, Singapore, April 24-28, 2025},
  publisher    = {OpenReview.net},
  year         = {2025},
  url          = {https://openreview.net/forum?id=din0lGfZFd},
  bibsource    = {dblp computer science bibliography, https://dblp.org}
}

@inproceedings{DBLP:conf/acl/ChenSZXSLWSW025,
  author       = {Yilong Chen and
                  Junyuan Shang and
                  Zhenyu Zhang and
                  Yanxi Xie and
                  Jiawei Sheng and
                  Tingwen Liu and
                  Shuohuan Wang and
                  Yu Sun and
                  Hua Wu and
                  Haifeng Wang},
  editor       = {Wanxiang Che and
                  Joyce Nabende and
                  Ekaterina Shutova and
                  Mohammad Taher Pilehvar},
  title        = {Inner Thinking Transformer: Leveraging Dynamic Depth Scaling to Foster
                  Adaptive Internal Thinking},
  booktitle    = {Proceedings of the 63rd Annual Meeting of the Association for Computational
                  Linguistics (Volume 1: Long Papers), {ACL} 2025, Vienna, Austria,
                  July 27 - August 1, 2025},
  pages        = {28241--28259},
  publisher    = {Association for Computational Linguistics},
  year         = {2025},
  url          = {https://aclanthology.org/2025.acl-long.1369/},
  bibsource    = {dblp computer science bibliography, https://dblp.org}
}

@inproceedings{DBLP:conf/naacl/PatelBG21,
  author       = {Arkil Patel and
                  Satwik Bhattamishra and
                  Navin Goyal},
  editor       = {Kristina Toutanova and
                  Anna Rumshisky and
                  Luke Zettlemoyer and
                  Dilek Hakkani{-}T{\"{u}}r and
                  Iz Beltagy and
                  Steven Bethard and
                  Ryan Cotterell and
                  Tanmoy Chakraborty and
                  Yichao Zhou},
  title        = {Are {NLP} Models really able to Solve Simple Math Word Problems?},
  booktitle    = {Proceedings of the 2021 Conference of the North American Chapter of
                  the Association for Computational Linguistics: Human Language Technologies,
                  {NAACL-HLT} 2021, Online, June 6-11, 2021},
  pages        = {2080--2094},
  publisher    = {Association for Computational Linguistics},
  year         = {2021},
  url          = {https://doi.org/10.18653/v1/2021.naacl-main.168},
  doi          = {10.18653/V1/2021.NAACL-MAIN.168},
  bibsource    = {dblp computer science bibliography, https://dblp.org}
}

@inproceedings{DBLP:conf/nips/HendrycksBKABTS21,
  author       = {Dan Hendrycks and
                  Collin Burns and
                  Saurav Kadavath and
                  Akul Arora and
                  Steven Basart and
                  Eric Tang and
                  Dawn Song and
                  Jacob Steinhardt},
  editor       = {Joaquin Vanschoren and
                  Sai{-}Kit Yeung},
  title        = {Measuring Mathematical Problem Solving With the {MATH} Dataset},
  booktitle    = {Proceedings of the Neural Information Processing Systems Track on
                  Datasets and Benchmarks 1, NeurIPS Datasets and Benchmarks 2021, December
                  2021, virtual},
  year         = {2021},
  url          = {https://datasets-benchmarks-proceedings.neurips.cc/paper/2021/hash/be83ab3ecd0db773eb2dc1b0a17836a1-Abstract-round2.html},
  bibsource    = {dblp computer science bibliography, https://dblp.org}
}

@article{DBLP:journals/corr/RoyR16,
  author       = {Subhro Roy and
                  Dan Roth},
  title        = {Solving General Arithmetic Word Problems},
  journal      = {CoRR},
  volume       = {abs/1608.01413},
  year         = {2016},
  url          = {http://arxiv.org/abs/1608.01413},
  eprinttype    = {arXiv},
  eprint       = {1608.01413},
  bibsource    = {dblp computer science bibliography, https://dblp.org}
}

@inproceedings{DBLP:conf/nips/DunefskyCN24,
  author       = {Jacob Dunefsky and
                  Philippe Chlenski and
                  Neel Nanda},
  editor       = {Amir Globersons and
                  Lester Mackey and
                  Danielle Belgrave and
                  Angela Fan and
                  Ulrich Paquet and
                  Jakub M. Tomczak and
                  Cheng Zhang},
  title        = {Transcoders find interpretable {LLM} feature circuits},
  booktitle    = {Advances in Neural Information Processing Systems 38: Annual Conference
                  on Neural Information Processing Systems 2024, NeurIPS 2024, Vancouver,
                  BC, Canada, December 10 - 15, 2024},
  year         = {2024},
  url          = {http://papers.nips.cc/paper\_files/paper/2024/hash/2b8f4db0464cc5b6e9d5e6bea4b9f308-Abstract-Conference.html},
  bibsource    = {dblp computer science bibliography, https://dblp.org}
}

@article{DBLP:journals/corr/abs-2507-06203,
  author       = {Ruijie Zhu and
                  Tianhao Peng and
                  Tianhao Cheng and
                  Xingwei Qu and
                  Jinfa Huang and
                  Dawei Zhu and
                  Hao Wang and
                  Kaiwen Xue and
                  Xuanliang Zhang and
                  Yong Shan and
                  Tianle Cai and
                  Taylor Kergan and
                  Assel Kembay and
                  Andrew Smith and
                  Chenghua Lin and
                  Binh Nguyen and
                  Yuqi Pan and
                  Yuhong Chou and
                  Zefan Cai and
                  Zhenhe Wu and
                  Yongchi Zhao and
                  Tianyu Liu and
                  Jian Yang and
                  Wangchunshu Zhou and
                  Chujie Zheng and
                  Chongxuan Li and
                  Yuyin Zhou and
                  Zhoujun Li and
                  Zhaoxiang Zhang and
                  Jiaheng Liu and
                  Ge Zhang and
                  Wenhao Huang and
                  Jason Eshraghian},
  title        = {A Survey on Latent Reasoning},
  journal      = {CoRR},
  volume       = {abs/2507.06203},
  year         = {2025},
  url          = {https://doi.org/10.48550/arXiv.2507.06203},
  doi          = {10.48550/ARXIV.2507.06203},
  eprinttype    = {arXiv},
  eprint       = {2507.06203},
  bibsource    = {dblp computer science bibliography, https://dblp.org}
}

@article{DBLP:journals/corr/abs-2412-16720,
  author       = {Aaron Jaech and
                  Adam Kalai and
                  Adam Lerer and
                  Adam Richardson and
                  Ahmed El{-}Kishky and
                  Aiden Low and
                  Alec Helyar and
                  Aleksander Madry and
                  Alex Beutel and
                  Alex Carney and
                  Alex Iftimie and
                  Alex Karpenko and
                  Alex Tachard Passos and
                  Alexander Neitz and
                  Alexander Prokofiev and
                  Alexander Wei and
                  Allison Tam and
                  Ally Bennett and
                  Ananya Kumar and
                  Andre Saraiva and
                  Andrea Vallone and
                  Andrew Duberstein and
                  Andrew Kondrich and
                  Andrey Mishchenko and
                  Andy Applebaum and
                  Angela Jiang and
                  Ashvin Nair and
                  Barret Zoph and
                  Behrooz Ghorbani and
                  Ben Rossen and
                  Benjamin Sokolowsky and
                  Boaz Barak and
                  Bob McGrew and
                  Borys Minaiev and
                  Botao Hao and
                  Bowen Baker and
                  Brandon Houghton and
                  Brandon McKinzie and
                  Brydon Eastman and
                  Camillo Lugaresi and
                  Cary Bassin and
                  Cary Hudson and
                  Chak Ming Li and
                  Charles de Bourcy and
                  Chelsea Voss and
                  Chen Shen and
                  Chong Zhang and
                  Chris Koch and
                  Chris Orsinger and
                  Christopher Hesse and
                  Claudia Fischer and
                  Clive Chan and
                  Dan Roberts and
                  Daniel Kappler and
                  Daniel Levy and
                  Daniel Selsam and
                  David Dohan and
                  David Farhi and
                  David Mely and
                  David Robinson and
                  Dimitris Tsipras and
                  Doug Li and
                  Dragos Oprica and
                  Eben Freeman and
                  Eddie Zhang and
                  Edmund Wong and
                  Elizabeth Proehl and
                  Enoch Cheung and
                  Eric Mitchell and
                  Eric Wallace and
                  Erik Ritter and
                  Evan Mays and
                  Fan Wang and
                  Felipe Petroski Such and
                  Filippo Raso and
                  Florencia Leoni and
                  Foivos Tsimpourlas and
                  Francis Song and
                  Fred von Lohmann and
                  Freddie Sulit and
                  Geoff Salmon and
                  Giambattista Parascandolo and
                  Gildas Chabot and
                  Grace Zhao and
                  Greg Brockman and
                  Guillaume Leclerc and
                  Hadi Salman and
                  Haiming Bao and
                  Hao Sheng and
                  Hart Andrin and
                  Hessam Bagherinezhad and
                  Hongyu Ren and
                  Hunter Lightman and
                  Hyung Won Chung and
                  Ian Kivlichan and
                  Ian O'Connell and
                  Ian Osband and
                  Ignasi Clavera Gilaberte and
                  Ilge Akkaya},
  title        = {OpenAI o1 System Card},
  journal      = {CoRR},
  volume       = {abs/2412.16720},
  year         = {2024},
  url          = {https://doi.org/10.48550/arXiv.2412.16720},
  doi          = {10.48550/ARXIV.2412.16720},
  eprinttype    = {arXiv},
  eprint       = {2412.16720},
  bibsource    = {dblp computer science bibliography, https://dblp.org}
}

@inproceedings{DBLP:conf/nips/Wei0SBIXCLZ22,
  author       = {Jason Wei and
                  Xuezhi Wang and
                  Dale Schuurmans and
                  Maarten Bosma and
                  Brian Ichter and
                  Fei Xia and
                  Ed H. Chi and
                  Quoc V. Le and
                  Denny Zhou},
  editor       = {Sanmi Koyejo and
                  S. Mohamed and
                  A. Agarwal and
                  Danielle Belgrave and
                  K. Cho and
                  A. Oh},
  title        = {Chain-of-Thought Prompting Elicits Reasoning in Large Language Models},
  booktitle    = {Advances in Neural Information Processing Systems 35: Annual Conference
                  on Neural Information Processing Systems 2022, NeurIPS 2022, New Orleans,
                  LA, USA, November 28 - December 9, 2022},
  year         = {2022},
  url          = {http://papers.nips.cc/paper\_files/paper/2022/hash/9d5609613524ecf4f15af0f7b31abca4-Abstract-Conference.html},
  bibsource    = {dblp computer science bibliography, https://dblp.org}
}

@article{DBLP:journals/corr/abs-2405-14838,
  author       = {Yuntian Deng and
                  Yejin Choi and
                  Stuart M. Shieber},
  title        = {From Explicit CoT to Implicit CoT: Learning to Internalize CoT Step
                  by Step},
  journal      = {CoRR},
  volume       = {abs/2405.14838},
  year         = {2024},
  url          = {https://doi.org/10.48550/arXiv.2405.14838},
  doi          = {10.48550/ARXIV.2405.14838},
  eprinttype    = {arXiv},
  eprint       = {2405.14838},
  bibsource    = {dblp computer science bibliography, https://dblp.org}
}

@article{DBLP:journals/corr/abs-2412-06769,
  author       = {Shibo Hao and
                  Sainbayar Sukhbaatar and
                  DiJia Su and
                  Xian Li and
                  Zhiting Hu and
                  Jason Weston and
                  Yuandong Tian},
  title        = {Training Large Language Models to Reason in a Continuous Latent Space},
  journal      = {CoRR},
  volume       = {abs/2412.06769},
  year         = {2024},
  url          = {https://doi.org/10.48550/arXiv.2412.06769},
  doi          = {10.48550/ARXIV.2412.06769},
  eprinttype    = {arXiv},
  eprint       = {2412.06769},
  bibsource    = {dblp computer science bibliography, https://dblp.org}
}

@article{DBLP:journals/corr/abs-2505-16552,
  author       = {Wenhui Tan and
                  Jiaze Li and
                  Jianzhong Ju and
                  Zhenbo Luo and
                  Jian Luan and
                  Ruihua Song},
  title        = {Think Silently, Think Fast: Dynamic Latent Compression of {LLM} Reasoning
                  Chains},
  journal      = {CoRR},
  volume       = {abs/2505.16552},
  year         = {2025},
  url          = {https://doi.org/10.48550/arXiv.2505.16552},
  doi          = {10.48550/ARXIV.2505.16552},
  eprinttype    = {arXiv},
  eprint       = {2505.16552},
  bibsource    = {dblp computer science bibliography, https://dblp.org}
}

@misc{bricken2023monosemanticity,
  title   = {Towards Monosemanticity: Decomposing Language Models With Dictionary Learning},
  author  = {Bricken, Trenton and Templeton, Adly and Batson, Joshua and Chen, Brian and Jermyn, Adam and Conerly, Tom and Turner, Nicholas L. and Anil, Cem and Denison, Carson and Askell, Amanda and Lasenby, Robert and Wu, Yifan and Kravec, Shauna and Schiefer, Nicholas and Maxwell, Tim and Joseph, Nicholas and Tamkin, Alex and Nguyen, Karina and McLean, Brayden and Burke, Josiah E and Hume, Tristan and Carter, Shan and Henighan, Tom and Olah, Chris},
  year    = {2023},
  howpublished = {\url{https://transformer-circuits.pub/2023/monosemantic-features/index.html}},
}

@article{DBLP:journals/corr/abs-2501-18823,
  author       = {Gon{\c{c}}alo Paulo and
                  Stepan Shabalin and
                  Nora Belrose},
  title        = {Transcoders Beat Sparse Autoencoders for Interpretability},
  journal      = {CoRR},
  volume       = {abs/2501.18823},
  year         = {2025},
  url          = {https://doi.org/10.48550/arXiv.2501.18823},
  doi          = {10.48550/ARXIV.2501.18823},
  eprinttype    = {arXiv},
  eprint       = {2501.18823},
  bibsource    = {dblp computer science bibliography, https://dblp.org}
}

@article{DBLP:journals/corr/abs-2502-21074,
  author       = {Zhenyi Shen and
                  Hanqi Yan and
                  Linhai Zhang and
                  Zhanghao Hu and
                  Yali Du and
                  Yulan He},
  title        = {{CODI:} Compressing Chain-of-Thought into Continuous Space via Self-Distillation},
  journal      = {CoRR},
  volume       = {abs/2502.21074},
  year         = {2025},
  url          = {https://doi.org/10.48550/arXiv.2502.21074},
  doi          = {10.48550/ARXIV.2502.21074},
  eprinttype    = {arXiv},
  eprint       = {2502.21074},
  bibsource    = {dblp computer science bibliography, https://dblp.org}
}

@article{DBLP:journals/corr/abs-2311-01460,
  author       = {Yuntian Deng and
                  Kiran Prasad and
                  Roland Fernandez and
                  Paul Smolensky and
                  Vishrav Chaudhary and
                  Stuart M. Shieber},
  title        = {Implicit Chain of Thought Reasoning via Knowledge Distillation},
  journal      = {CoRR},
  volume       = {abs/2311.01460},
  year         = {2023},
  url          = {https://doi.org/10.48550/arXiv.2311.01460},
  doi          = {10.48550/ARXIV.2311.01460},
  eprinttype    = {arXiv},
  eprint       = {2311.01460},
  bibsource    = {dblp computer science bibliography, https://dblp.org}
}

@article{DBLP:journals/corr/abs-2501-04682,
  author       = {Violet Xiang and
                  Charlie Snell and
                  Kanishk Gandhi and
                  Alon Albalak and
                  Anikait Singh and
                  Chase Blagden and
                  Duy Phung and
                  Rafael Rafailov and
                  Nathan Lile and
                  Dakota Mahan and
                  Louis Castricato and
                  Jan{-}Philipp Fr{\"{a}}nken and
                  Nick Haber and
                  Chelsea Finn},
  title        = {Towards System 2 Reasoning in LLMs: Learning How to Think With Meta
                  Chain-of-Thought},
  journal      = {CoRR},
  volume       = {abs/2501.04682},
  year         = {2025},
  url          = {https://doi.org/10.48550/arXiv.2501.04682},
  doi          = {10.48550/ARXIV.2501.04682},
  eprinttype    = {arXiv},
  eprint       = {2501.04682},
  bibsource    = {dblp computer science bibliography, https://dblp.org}
}

@article{DBLP:journals/corr/abs-2407-21783,
  author       = {Llama Team},
  title        = {The Llama 3 Herd of Models},
  journal      = {CoRR},
  volume       = {abs/2407.21783},
  year         = {2024},
  url          = {https://doi.org/10.48550/arXiv.2407.21783},
  doi          = {10.48550/ARXIV.2407.21783},
  eprinttype    = {arXiv},
  eprint       = {2407.21783},
  bibsource    = {dblp computer science bibliography, https://dblp.org}
}

@article{DBLP:journals/corr/abs-2501-12948,
  author       = {DeepSeek{-}AI},
  title        = {DeepSeek-R1: Incentivizing Reasoning Capability in LLMs via Reinforcement
                  Learning},
  journal      = {CoRR},
  volume       = {abs/2501.12948},
  year         = {2025},
  url          = {https://doi.org/10.48550/arXiv.2501.12948},
  doi          = {10.48550/ARXIV.2501.12948},
  eprinttype    = {arXiv},
  eprint       = {2501.12948},
  bibsource    = {dblp computer science bibliography, https://dblp.org}
}

@article{DBLP:journals/tacl/GevaKSKRB21,
  author       = {Mor Geva and
                  Daniel Khashabi and
                  Elad Segal and
                  Tushar Khot and
                  Dan Roth and
                  Jonathan Berant},
  title        = {Did Aristotle Use a Laptop? {A} Question Answering Benchmark with
                  Implicit Reasoning Strategies},
  journal      = {Trans. Assoc. Comput. Linguistics},
  volume       = {9},
  pages        = {346--361},
  year         = {2021},
  url          = {https://doi.org/10.1162/tacl\_a\_00370},
  doi          = {10.1162/TACL\_A\_00370},
  bibsource    = {dblp computer science bibliography, https://dblp.org}
}

@inproceedings{knowprompt,
  author       = {Xiang Chen and
                  Ningyu Zhang and
                  Xin Xie and
                  Shumin Deng and
                  Yunzhi Yao and
                  Chuanqi Tan and
                  Fei Huang and
                  Luo Si and
                  Huajun Chen},
  editor       = {Fr{\'{e}}d{\'{e}}rique Laforest and
                  Rapha{\"{e}}l Troncy and
                  Elena Simperl and
                  Deepak Agarwal and
                  Aristides Gionis and
                  Ivan Herman and
                  Lionel M{\'{e}}dini},
  title        = {KnowPrompt: Knowledge-aware Prompt-tuning with Synergistic Optimization
                  for Relation Extraction},
  booktitle    = {{WWW} '22: The {ACM} Web Conference 2022, Virtual Event, Lyon, France,
                  April 25 - 29, 2022},
  pages        = {2778--2788},
  publisher    = {{ACM}},
  year         = {2022},
  url          = {https://doi.org/10.1145/3485447.3511998},
  doi          = {10.1145/3485447.3511998},
  bibsource    = {dblp computer science bibliography, https://dblp.org}
}

@inproceedings{deco,
  author       = {Chenxi Wang and
                  Xiang Chen and
                  Ningyu Zhang and
                  Bozhong Tian and
                  Haoming Xu and
                  Shumin Deng and
                  Huajun Chen},
  title        = {{MLLM} can see? Dynamic Correction Decoding for Hallucination Mitigation},
  booktitle    = {The Thirteenth International Conference on Learning Representations,
                  {ICLR} 2025, Singapore, April 24-28, 2025},
  publisher    = {OpenReview.net},
  year         = {2025},
  url          = {https://openreview.net/forum?id=4z3IguA4Zg},
  bibsource    = {dblp computer science bibliography, https://dblp.org}
}

@ARTICLE{11193717,
  author={Chen, Xiang and Ou, Yixin and Feng, Quan and Li, Lei and Li, Piji and Ye, Haibo and Huang, Sheng-Jun and Qiao, Shuofei and Deng, Shumin and Chen, Huajun and Zhang, Ningyu},
  journal={IEEE Transactions on Audio, Speech and Language Processing}, 
  title={Retrieval-Augmented Prompt Learning for Pre-Trained Foundation Models}, 
  year={2025},
  volume={33},
  number={},
  pages={4311-4322},
  doi={10.1109/TASLPRO.2025.3608936}}

\newpage
\appendix

\section{Implementation Details}
\label{app:impl}

\paragraph{Model Hyperparameters}
All experiments are conducted using the \texttt{Llama-3.2-1B-Instruct} backbone~\cite{DBLP:journals/corr/abs-2407-21783}. The backbone parameters are frozen throughout training and adapted using LoRA, while the LTT parameters are trained directly. We set the LoRA rank to $r_{\text{LoRA}}=128$ and the scaling factor to $\alpha_{\text{LoRA}}=32$ for all experiments. The latent reasoning module adopts the LTT architecture defined in Section~\ref{sec:ltt}, where the latent feature dimension is set to $d_{\text{feat}} = 16 \times d_{\text{model}}$. Unless otherwise specified, the sparsity budget is fixed to $k=128$ active features per latent step. The LTT encoder is initialized with orthogonal weights, while decoder columns are constrained to unit norm to stabilize feature scaling. The decoder bias is initialized using the empirical mean hidden state computed over the training corpus. The skip path linear adapter is initialized to zero, ensuring that non-linear transition residuals are initially routed through the sparse path. All latent activations use ReLU nonlinearity, and the latent temperature is fixed to $1.0$.

\paragraph{Training and Optimization Protocol}
All models are trained exclusively using \textbf{Supervised Fine-Tuning (SFT)}; we do not employ reinforcement learning or policy optimization in any experiment. We utilize the AdamW optimizer with a base learning rate of $1 \times 10^{-4}$ and weight decay of $1 \times 10^{-2}$. To accommodate the different convergence properties of the backbone and the sparse transcoder, we implement a hierarchical learning rate strategy:
\begin{table*}[t]
\centering
\small
\begin{tabular}{lccc}
\toprule
\textbf{Parameter Group} & \textbf{Learning Rate} & \textbf{Weight Decay} & \textbf{Functional Role} \\
\midrule
Backbone (LoRA) & $1 \times 10^{-4}$ & 0.1 & Backbone adaptation \\
Skip Path ($\mathbf{W}_{\text{skip}}$) & $2 \times 10^{-4}$ & 0.0 & Skip path \\
LTT Encoder & $2 \times 10^{-4}$ & 0.0 & Sparse path encoder \\
LTT Decoder & $2 \times 10^{-4}$ & 0.0 & Sparse path decoder \\
\bottomrule
\end{tabular}
\end{table*}

The sparsity budget is fixed at $k=128$ throughout training. Gradient clipping with a maximum norm of $1.0$ is applied to stabilize training. All experiments are conducted on \textbf{two NVIDIA RTX PRO6000 GPUs} using mixed-precision training. Unless otherwise stated, results are averaged over five random seeds (0--4).

 \section{Sensitivity Analysis of Compression Ratio \texorpdfstring{$r$}{r}}
\label{app:r_sensitivity}

\begin{figure}[t]
\centering
\includegraphics[width=\columnwidth]{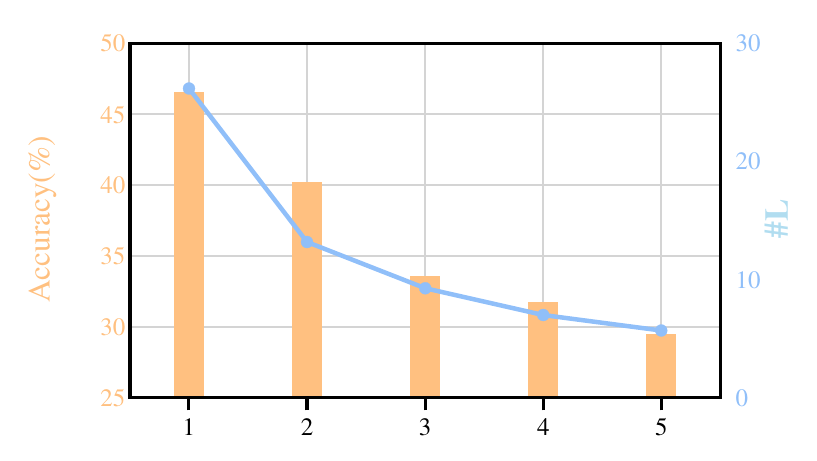}
\caption{\textbf{Accuracy and reasoning length (\#L) across varying compression ratios $r$}. The orange bars denote reasoning accuracy (\%) , while the blue line represents the average number of latent steps (\#L) per problem. Increasing $r$ leads to significantly more compact trajectories but results in a corresponding decline in accuracy.}
\label{fig:r_tradeoff}
\end{figure}

In the main paper, we fix the sparsity budget to $k=128$ and focus on comparing established latent compression ratios. In this appendix, we provide supplementary results analyzing how the compression strength $r$ affects reasoning performance and trajectory compactness. Specifically, we evaluate \ours{} models across a range of compression ratios $r \in \{1, 2, 3, 4, 5\}$, while maintaining a fixed sparsity budget of $k=128$ and keeping all other hyperparameters constant. Figure~\ref{fig:r_tradeoff} illustrates the corresponding accuracy and average latent reasoning length. As the compression ratio $r$ increases from 1 to 5 , the latent trajectory becomes progressively shorter, resulting in substantially and monotonically reduced reasoning length. However, aggressive compression also leads to a gradual degradation in accuracy, reflecting a reduced capacity of the latent manifold to preserve fine-grained reasoning signals under aggressive compression. These results demonstrate a clear and smooth accuracy–efficiency Pareto frontier controlled by the compression ratio $r$, confirming that \ours{} enables controllable latent reasoning compression without requiring architectural changes or additional model retraining.

\section{Training and Inference Overhead}
\label{app:overhead}

We analyze the computational overhead of \ours{} relative to explicit CoT and the dense \textit{CoLaR} baseline. All measurements were conducted on a single NVIDIA RTX PRO6000 GPU using FP16 precision with a batch size of 1.

\begin{table*}[t]
\centering
\small
\caption{Efficiency profiling on GSM8k-Aug. Per-step latency and Peak VRAM were measured during inference. End-to-end (E2E) time is calculated as Average Steps $\times$ Step Latency.}
\label{tab:overhead_detailed}
\vspace{2mm}
\begin{tabular}{l ccc cc}
\toprule
\textbf{Method} & \textbf{Acc. (\%)} & \textbf{Avg. Steps (\#L)} & \textbf{Step Latency} & \textbf{Policy Overhead} & \textbf{Peak VRAM} \\
\midrule
CoT (Explicit)  & \textbf{49.4} & 25.60 & 9.26 ms & --- & \textbf{5.14 GB} \\
\midrule
CoLaR-2 (Dense) & 40.1 & \textbf{12.70} & \textbf{8.98 ms} & 1.69\% & 5.21 GB \\
\rowcolor{gray!20}
\ours-2 (Ours)  & 40.2 & 13.18 & 10.66 ms & 10.53\% & 5.66 GB \\
\bottomrule
\end{tabular}
\end{table*}

\paragraph{Inference Latency.} As shown in Table~\ref{tab:overhead_detailed}, \ours{} introduces additional computation at each latent step due to the $16\times$ feature expansion in the LTT encoder and the $\mathrm{Top}\text{-}k$ selection process. Specifically, the \textbf{1.02 ms} required for the LTT operation results in a per-step latency of \textbf{10.66 ms}, representing a \textbf{1.68 ms} increase over the dense baseline (8.98 ms). However, this per-step increase is effectively offset by the reduction in total reasoning length. Compared to explicit CoT, \ours-2 reduces the reasoning sequence from \textbf{25.6} tokens to \textbf{13.18} steps. Consequently, the total end-to-end (E2E) inference time for a representative reasoning chain is reduced from \textbf{237.1 ms} (CoT) to \textbf{140.5 ms} (\ours-2), achieving a \textbf{40.7\%} speed improvement. This indicates that \ours{} improves efficiency primarily by significantly reducing reasoning length rather than minimizing per-step computation.

\paragraph{Memory and Training.} During training, the backbone remains frozen, and overhead arises primarily from sparse feature selection and latent trajectory supervision. The introduction of \textbf{138.45M} parameters in the LTT module increases the peak VRAM usage by approximately \textbf{450 MB} compared to the dense model (5.66 GB vs. 5.21 GB) and \textbf{520 MB} compared to the CoT baseline. While the static memory footprint is higher, the significantly shorter latent trajectories lead to a slower growth of the \textbf{KV Cache} during generation. This makes sparse latent reasoning a practical and scalable alternative to explicit CoT, as the benefits of trajectory compression outweigh the marginal increases in per-step latency and static parameter storage.

\section{Cross-Backbone Consistency}
\label{app:harder_tasks}

To evaluate whether sparse latent reasoning extends beyond small-scale backbones, we conduct scalability experiments with \ours{} on \Deepseek~\cite{DBLP:journals/corr/abs-2501-12948} and \Llama~\cite{DBLP:journals/corr/abs-2407-21783}. All experiments are performed on the MATH benchmark, which constitutes a substantially more challenging reasoning setting due to its complex symbolic dependencies and multi-step deduction requirements. For all scalability evaluations, \ours{} adopts a fixed configuration with a compression ratio of $r=2$ and a sparsity budget of $k=128$. This setting follows the training protocol introduced in Section~4.1 and enables a direct comparison with the CoLaR baseline under the same compression ratio.

Across both DeepSeek-based and Llama-based backbones, \ours{} consistently produces substantially shorter reasoning trajectories than explicit CoT decoding. The compression advantage remains stable as model scale increases, indicating that sparse latent transitions can internalize nontrivial reasoning procedures even within high-dimensional representation spaces. Although a performance gap relative to CoT remains, the gap is expected given the difficulty of preserving fine-grained symbolic reasoning within a fully latent reasoning process. Importantly, the overall behavior of \ours{} remains highly consistent across the two model families. Both accuracy and reasoning-length trends exhibit similar patterns for the DeepSeek and Llama backbones despite differences in architecture design and distillation strategies. This stability across heterogeneous backbones suggests that the effectiveness of sparse latent reasoning primarily originates from the LTT formulation rather than from backbone-specific properties. Overall, the results demonstrate that \ours{} provides a scalable and structurally consistent framework for efficient reasoning in LLMs while preserving the efficiency--interpretability trade-off under increasing model capacity.

\begin{table*}[t]
    \centering
    \small
    \caption{Scalability evaluation on 7B+ backbones over the MATH benchmark. All models use a compression ratio of $r=2$ and a sparsity budget of $k=128$. Across different model families, \ours{} preserves a stable efficiency--accuracy trade-off, indicating that the sparse transition formulation can be applied across model families, while hard symbolic reasoning remains challenging under latent compression.}
    \label{tab:multi_model_results}
    \vspace{2mm}
    \renewcommand{\arraystretch}{1.2}
    \setlength{\tabcolsep}{18pt}
    \begin{tabular}{l cc cc}
        \toprule
        \multirow{2}{*}{\textbf{Method}} & \multicolumn{2}{c}{\textbf{DeepSeek-R1-Distill-Qwen-7B}} & \multicolumn{2}{c}{\textbf{Llama-3.2-8B-Instruct}} \\
        \cmidrule(lr){2-3} \cmidrule(lr){4-5}
        & Acc. & \# L & Acc. & \# L \\
        \midrule
        CoT & $35.2_{\pm .31}$ & $198.0_{\pm 1.2}$ & $24.1_{\pm .27}$ & $200.0_{\pm 1.5}$ \\
        CoLaR & $16.5_{\pm .14}$ & $81.2_{\pm .23}$ & $8.0_{\pm .19}$ & $85.4_{\pm .44}$ \\
        \midrule
        \rowcolor{gray!10}
        \ours{} & $15.9_{\pm .11}$ & $72.9_{\pm .00}$ & $10.9_{\pm .23}$ & $82.9_{\pm .01}$ \\
        \bottomrule
    \end{tabular}
\end{table*}

\begin{figure*}[t]
    \centering
    \begin{subfigure}[b]{0.48\textwidth}
        \centering
        \includegraphics[width=\textwidth]{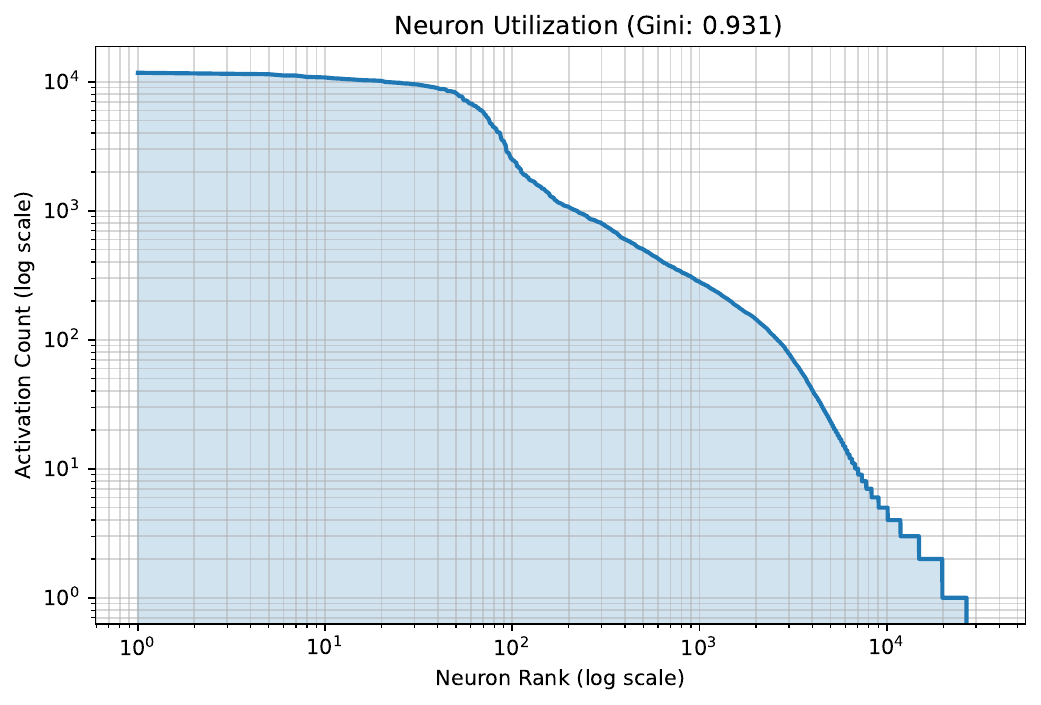}
        \caption{\ours{}}
        \label{fig:util_full}
    \end{subfigure}
    \hfill
    \begin{subfigure}[b]{0.48\textwidth}
        \centering
        \includegraphics[width=\textwidth]{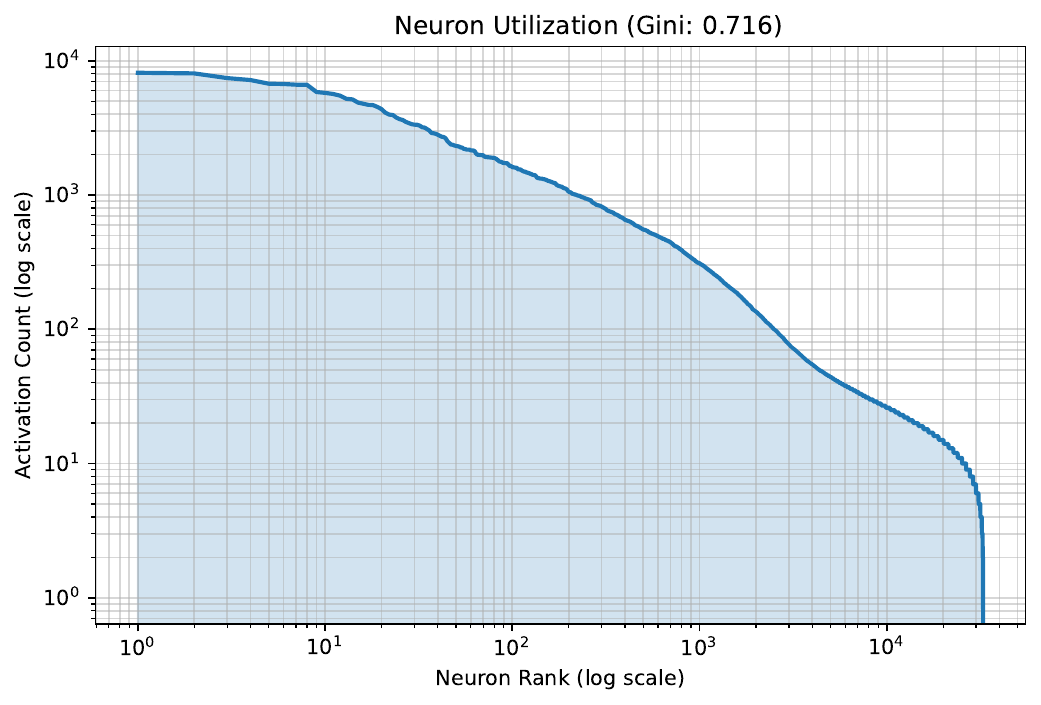}
        \caption{\ours{} without skip path}
        \label{fig:util_sae_only}
    \end{subfigure}
    \caption{Feature utilization under Top-$k$ inference. The full model shows a more concentrated utilization pattern, while the model without the skip path uses features more diffusely.}
    \label{fig:Utilization}
\end{figure*}

\section{Latent Structural Analysis}
\label{app:latent_structure}

To comprehensively understand the internal dynamics of sparse latent reasoning, we analyze the structural properties of \ours{} at both the micro level of individual feature activations and the macro level of global trajectory evolution.

\paragraph{Feature Utilization} 
At the micro level, we analyze the utilization frequency of LTT features by counting their activations under $\mathrm{Top}\text{-}k$ inference. As shown in Figure~\ref{fig:Utilization}, the resulting rank–frequency distribution follows a clear power-law pattern. In the full \ours{} architecture, feature utilization is highly concentrated, yielding a Gini coefficient of approximately $0.93$, which indicates the strong specialization of a small subset of latent features. In contrast, when the skip path is removed and only the sparse path is retained, feature utilization becomes significantly more diffuse, reducing the Gini coefficient to $0.71$. This comparison highlights the crucial functional role of the skip path: by absorbing predictable background drift, it allows sparse features to highly specialize in discrete semantic innovations.

\begin{figure*}[t]
    \centering
    \includegraphics[width=0.98\linewidth]{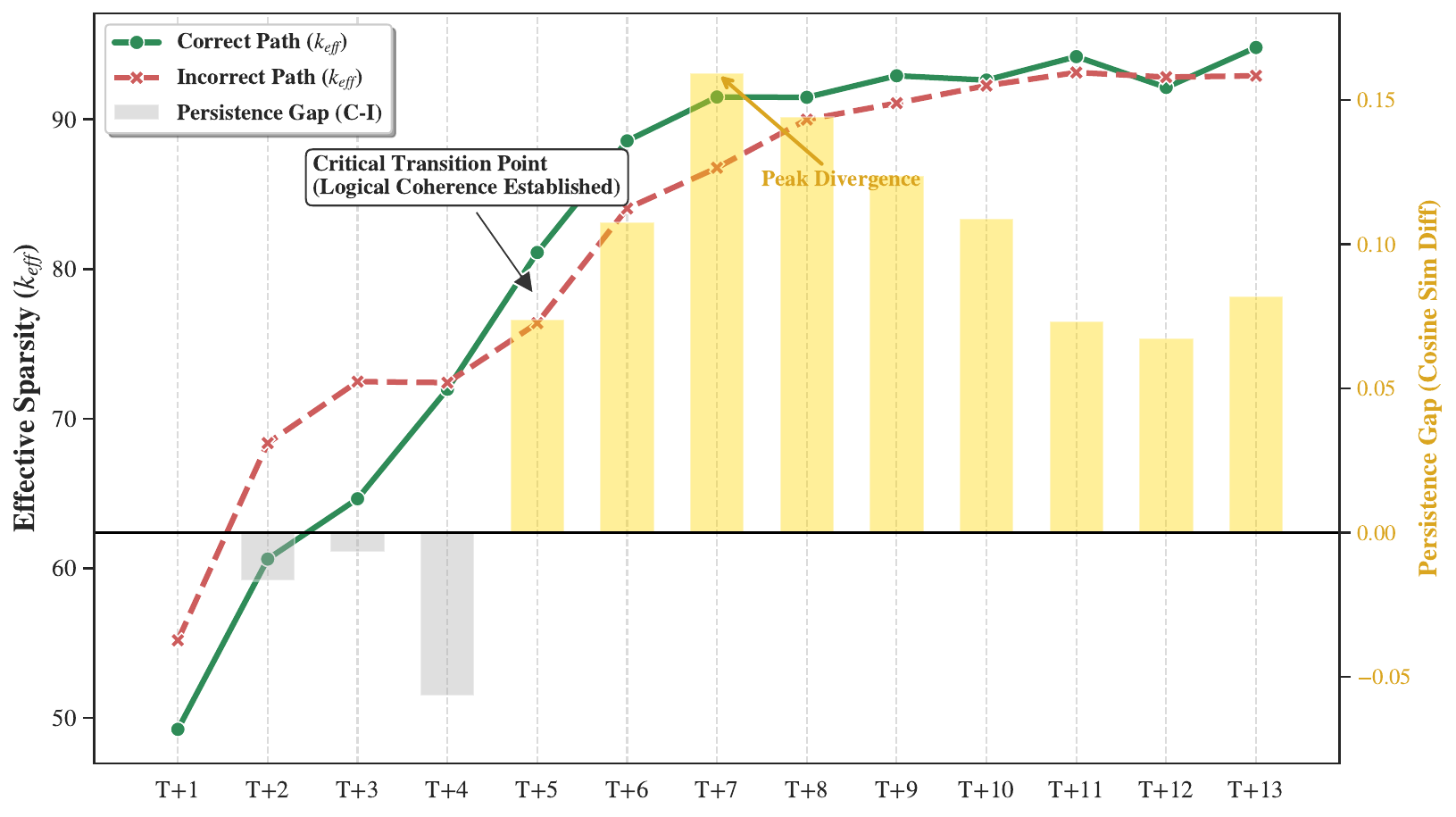}
    \caption{Step-wise effective sparsity $k_{\mathrm{eff}}$ and feature persistence along latent reasoning trajectories. The model is trained with $r=2$ and $k=128$; sparsity is fixed at inference time. Correct and incorrect trajectories are aggregated separately.}
    \label{fig:latent_mechanism}
\end{figure*}

\begin{figure}[t]
    \centering
    \includegraphics[width=\columnwidth]{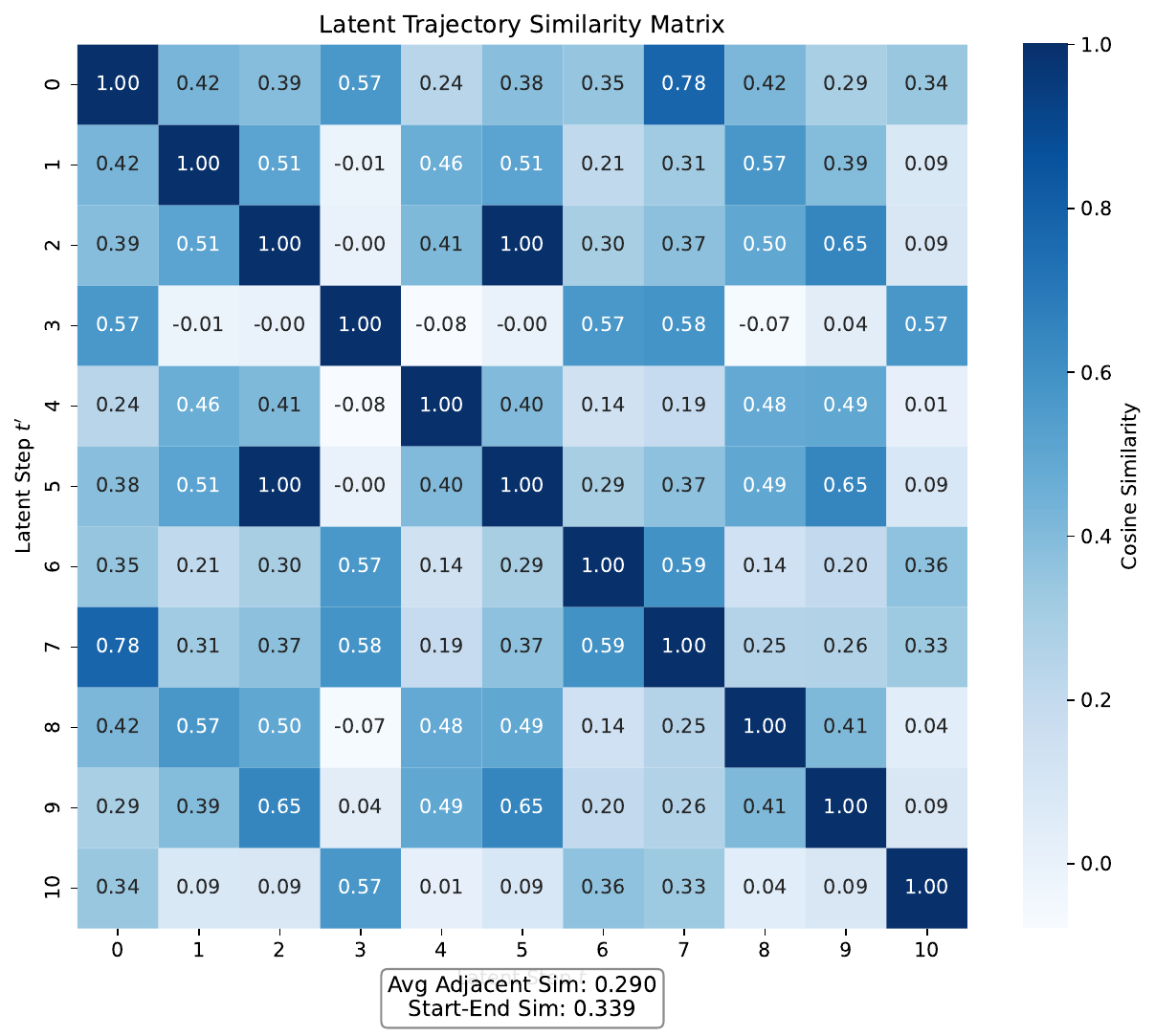}
    \caption{Cosine similarity between reconstructed latent embeddings across latent steps. High similarity near the diagonal indicates local temporal continuity, while lower similarity between early and late steps reflects gradual latent-state change.}
    \label{fig:Trajectory}
\end{figure}

\paragraph{Latent Trajectory Coherence}
Given the highly specialized and sparse nature of these individual updates, a natural question is whether such hard representational bottlenecks disrupt the overall continuity of the reasoning process. To address this, we examine the macro-level semantic evolution of latent embeddings across successive steps. Figure~\ref{fig:Trajectory} illustrates the cosine similarity matrix between reconstructed latent embeddings within a single reasoning trajectory. Adjacent latent steps exhibit high similarity, which preserves local temporal continuity. Meanwhile, as illustrated in Figure~\ref{fig:latent_mechanism}, the similarity between the initial and final steps is substantially lower (Start–End similarity: $0.339$), indicating a smooth yet progressive semantic drift. This pattern confirms that latent reasoning in \ours{} follows a coherent, goal-directed semantic trajectory, successfully avoiding both the collapse into static representations and the instability of random fluctuations.

\end{document}